\pdfoutput=1
\documentclass[10pt,twocolumn,letterpaper]{article}
\usepackage[pagenumbers]{cvpr}
\usepackage{times}
\usepackage{soul}
\usepackage{url}
\usepackage[utf8]{inputenc}
\usepackage{graphicx}
\usepackage{float}
\usepackage{amsmath}
\usepackage{amssymb}
\usepackage{amsthm}
\usepackage{booktabs}
\usepackage{algorithm}
\usepackage{algorithmic}
\usepackage{multirow}
\usepackage{xcolor}

\definecolor{cvprblue}{rgb}{0.21,0.49,0.74}
\usepackage[breaklinks,colorlinks,allcolors=cvprblue]{hyperref}
\newtheorem{lemma}{Lemma}
\title{SNN-STEM: Towards High-Performance Aerial Object Detection with Frequency-Enhanced Spiking Neural Networks}
\author{%
Yuxuan Hou\textsuperscript{1,*}\quad
Wentao Tong\textsuperscript{2,*}\quad
Di Yu\textsuperscript{2}\quad
Linshan Jiang\textsuperscript{1}\quad
Xin Du\textsuperscript{2}\\[3pt]
\textsuperscript{1}Southern University of Science and Technology\qquad
\textsuperscript{2}Zhejiang University\\[2pt]
{\small \textsuperscript{*}Equal contribution.}%
}
\hypersetup{pdftitle={SNN-STEM: Towards High-Performance Aerial Object Detection with Frequency-Enhanced Spiking Neural Networks},pdfauthor={Yuxuan Hou, Wentao Tong, Di Yu, Linshan Jiang, Xin Du}}
\begin{document}
\maketitle
\begin{abstract}
    Object detection from unmanned aerial vehicles (UAVs) is an important task in edge intelligence, but UAVs' endurance is severely limited by the high energy consumption of conventional Artificial Neural Networks (ANNs). While Spiking Neural Networks (SNNs) offer a promising, energy-efficient alternative, an accuracy gap remains in complex aerial scenes. In this paper, we identify that this gap may derive from the intrinsic frequency-domain properties of spiking neurons: they act as a temporal low-pass filter, attenuating high-frequency information crucial for encoding precision, and as a spatial ``low-cut'' filter, eroding low-frequency components vital for representing prevalent small objects in aerial imagery. To overcome these challenges, we propose SNN-STEM: spatial and temporal enhancement modules. Our temporal module enriches input with multi-frequency cues via a novel RGB-F encoding, while our spatial module supplements attenuated low-frequency signals using a frequency-aware shortcut. Extensive experiments on the VisDrone and VHR-10 datasets demonstrate that SNN-STEM consistently boosts the performance of state-of-the-art SNN detectors with minimal overheads. In particular, the enhanced SpikeYOLO models improve both mAP50 and mAP50:95 on VHR-10, supporting the effectiveness of SNN-STEM for high-performance, energy-efficient aerial object detection.
\end{abstract}

\section{Introduction}
Benefiting from the flourishing global drone industry, unmanned aerial vehicles (UAVs)~\cite{uav} have been applied in many areas such as security and surveillance~\cite{uav,visdrone_tpami2022}, search and rescue, and sports analysis. Essentially, object detection is one of the most studied tasks when UAVs are being used in those applications~\cite{uavob-survey}. Most object detection frameworks based on YOLO~\cite{yolo,yolo9000_2017,darknet} and RCNN~\cite{rcnn,fasterrcnn2015} use artificial neural networks (ANNs), which have high computational complexity and energy consumption. Other representative ANN detectors include SSD~\cite{ssd2016} and RetinaNet~\cite{focal2017}, while feature pyramids support multi-scale detection~\cite{fpn2017}. However, small aerial vehicles usually carry small devices with small battery capacity, further limiting their endurance when ANN-based object detection applications are running on the device. As indicated in~\cite{uavenduration}, when an onboard computer system (e.g., DJI Manifold 2-G) performs a real-time object recognition task~\cite{uavtask}, the UAV’s flight endurance is often reduced to less than 20 minutes. Thus, the severe energy limitation demands the adoption of a more energy-efficient neural network paradigm on UAVs.

Spiking Neural Networks (SNNs), known as the third generation of neural networks~\cite{maass1997}, potentially serve as a more efficient and biologically inspired way to perform neural network-based recognition tasks on UAVs~\cite{snn}. Their energy efficiency (SNNs consume up to 16.2$\times$ less energy than ANNs~\cite{uav-snn}) and practicality when being deployed on UAVs for small object detection tasks~\cite{uav-snn}, event-camera-based detection tasks~\cite{uav-dvs1,uav-dvs2}, and UAV swarm tasks~\cite{uav-swarm} have been demonstrated. However, SNNs using spiking neurons as activation functions have a very different working mechanism compared to ANNs~\cite{annsnngap}, raising concerns about some unique challenges when SNNs are used to handle complex aerial object detection tasks.

Normally, aerial images captured by UAVs are often characterized by the prevalence of numerous small objects~\cite{SODA,visdrone_tpami2022}. In a 1024×1024 image, for instance, some objects may occupy less than 10 pixels~\cite{visdrone}. Aerial benchmarks also exhibit substantial scale variation~\cite{dota2018}, and clustered detection~\cite{clusdet2019} and sparse high-resolution querying~\cite{querydet2022} address complementary aspects of these scenes. Such characteristics may pose unique challenges when applying SNNs to aerial object detection, which mainly arise from SNNs' defining feature compared to ANNs, i.e., the spiking neuron. Compared to conventional activation functions like ReLU~\cite{nair2010} and SiLU~\cite{silu}, spiking neurons exhibit far more complicated dynamics, two of which attract most of our attention in this paper's discussion, i.e., temporal integration of membrane potential and the steep Heaviside activation function. With the temporal integration, as indicated in~\cite{maxformer}, the leaky integrate-and-fire (LIF) neuron is proved to be a \textbf{temporal low-pass filter} in their discussion and tends to filter out temporal high-frequency components of the signals, degrading the accuracy of image classification. Motivated by that, we think further about spatial domain and find that LIF neuron with Heaviside function can significantly degrade \textbf{spatial} low-frequency information. These two properties of SNNs could bring huge impacts to the detection of aerial images. Firstly, the encoding precision of SNNs' temporal information degrades when high-frequency components are lost. This information, defined by the binary spike encoding of floating-point values~\cite{rueckauer2017,optimal2021}, relies on these high-frequency components to generate accurate predictions for precision-critical output (e.g., bounding boxes). Secondly, the unwarranted loss of low-frequency spatial information may turn those small objects into noise, thus omitting them during inference.

We therefore propose dedicated solutions to overcome those challenges, thereby enabling high-performance aerial object detection with SNNs on UAVs. \textbf{Temporally}, we enhance the input representation. Instead of simply repeating the same image, we apply a wavelet transform~\cite{dwt,mallat1989} to decompose the image into multiple frequency subbands. These subbands are then concatenated with the original image from high to low frequencies, forming an RGB-F input. \textbf{Spatially}, we directly supplement the diminished low-frequency components. Building upon a standard residual block~\cite{resnet2016,identity2016}, we introduce a low-frequency shortcut. This branch utilizes average pooling and Fast Fourier Convolution (FFC)~\cite{ffc2020} to extract and reinforce low-frequency signals, which are then added to the membrane potential to preserve essential spatial information. To support our findings and demonstrate the effectiveness of our proposed modules, we conduct extensive experiments on two widely-used SNN architectures for object detection, i.e., the EMS-YOLO pipeline~\cite{ems-yolo} and the SpikeYOLO pipeline~\cite{SpikeYOLO}. Intriguingly, with a minimal increase in parameters, our spatial and temporal enhancement modules boost the mAP50 and mAP50:95 of SpikeYOLO baselines by an average of 2.8 and 2.0 percentage points (pp), respectively, on the VHR-10 dataset~\cite{vhr-10}. On the more challenging VisDrone dataset~\cite{visdrone}, the enhanced SpikeYOLO models maintain an average lead of 1.2~pp in mAP50 and 0.6~pp in mAP50:95 over the baselines.
Our contributions can be summarized as follows:
\begin{itemize}
    \item As far as we can see, this paper is the first to systematically discuss unique challenges when applying SNNs on difficult aerial object detection tasks (Section~\ref{sec:mot}). Based on our discussions, we propose SNN-STEM, \underline{\textbf{s}}patial and \underline{\textbf{t}}emporal \underline{\textbf{e}}nhancement \underline{\textbf{m}}odules to tackle spatial and temporal challenges of SNNs, respectively.
    \item Temporally, we propose an input enhancement module using discrete wavelet transform (DWT) to inject rich temporal information into the input, compensating for suppressed temporal high-frequency information. 
    \item Spatially, we develop a simplified Fast Fourier Convolution (FFC) module, which is used in an additional AvgPool-FFC shortcut path in residual blocks, to precisely capture and repair spatial low-frequency information that is filtered out in the main path. 
    \item We conduct extensive experiments to validate our approach. On the VisDrone dataset, the enhanced EMS-YOLO and SpikeYOLO pipelines achieve average mAP50 improvements of 1.1~pp and 1.2~pp, respectively. On the VHR-10 dataset, the improvements are 1.5~pp and 2.8~pp, respectively.
\end{itemize}

\section{Related Work}
\subsection{Spatial-Temporal Analysis for SNNs}
Modern Spiking Neural Networks (SNNs) often adopt structures similar to ANNs (e.g., MS-ResNet~\cite{ms-resnet}, SpikFormer~\cite{spikformer2023}), yet they still exhibit a noticeable performance gap. To understand this discrepancy, research has increasingly focused on the spiking neuron itself, whose mechanism differs fundamentally from the ReLU activation used in ANNs. Related architectures include Spike-driven Transformer~\cite{spikformer}, Meta-SpikeFormer~\cite{sdtv2_2024}, and SpikingResformer~\cite{spikingresformer2024}. Neuron-level studies model diverse firing behaviors~\cite{izhikevich}, learn membrane time constants~\cite{plif}, combine gated dynamics~\cite{glif2022}, and introduce complementary memory~\cite{clif}, and parallel spiking neurons further restructure temporal dependencies~\cite{psn2023}. 

Li et al.~\cite{temporal2} indicate that an SNN can approximate an ANN given enough time steps, implying an important temporal property that a spiking neuron is able to encode floating activations of ReLU. However, they do not explain the gap between ANNs and SNNs with few time steps. Rombouts et al.~\cite{fpsn} study fractional predictive spike encoding and temporal signal filtering. In contrast, SWformer~\cite{swt} explains the property of the LIF neuron as ``struggling to learn high frequency patterns''. Recently, MaxFormer~\cite{maxformer} revealed the nature of the LIF neuron theoretically, explaining it as a temporal low-pass filter. Frequency-domain visual representations have also been explored through frequency-channel input learning~\cite{freqlearn2020}, multi-spectral channel attention~\cite{fcanet2021}, and global filtering~\cite{gfnet2021}. Wavelet-based networks use multiresolution decomposition for noise robustness~\cite{wavecnet2020}, attention~\cite{wavevit2022}, and large receptive fields~\cite{wtconv2024}, while Octave Convolution separates high- and low-frequency feature groups~\cite{octconv2019}. Existing studies often examine spiking neurons in isolation (i.e., either temporally or spatially), and their biased evaluations motivate us to further explore spatial and temporal mechanisms of SNNs together.

\subsection{SNNs for Object Detection}
Early SNN-based object detection models relied on ANN-SNN conversion~\cite{spiking-yolo,spike-calib}, which requires many time steps to match ANN performance, resulting in high memory costs and latency. Related conversion studies address residual membrane potentials~\cite{rmp2020} and optimal conversion conditions~\cite{optimal2021}. This limits their practicality in real-time applications such as autonomous vehicles and UAVs. Therefore, directly-trained SNNs that need very few time steps during inference are becoming the main trend in energy-efficient edge intelligence. Direct training builds on surrogate-gradient optimization~\cite{sg} and spatio-temporal backpropagation~\cite{stbp2018,direct2019}, with normalization~\cite{tdbn2021} and temporal reweighting~\cite{tet2022} improving optimization. Differentiation on spike representations~\cite{dsr2022} and hybrid conversion with spike-timing backpropagation~\cite{hybrid2020} provide further routes to the accuracy-latency trade-off. Su et al. introduced EMS-YOLO~\cite{ems-yolo}, a lightweight pipeline based on the Tiny-YOLO architecture~\cite{yolov3_2018,ems-yolo}. While it requires only 5 time steps for inference, its performance lags significantly behind comparable ANN models due to the use of standard LIF spiking neurons. SpikeYOLO~\cite{SpikeYOLO} then proposed I-LIF, which produces quantized integers as spikes instead of 0-1 spikes, to improve the performance of SNNs. Moreover, Miao et al. proposed an ANN-SNN hybrid model, SpikingYOLOX~\cite{spikingyolox,yolox2021}, utilizing darknet~\cite{yolov3_2018} as the backbone and integrating fast Fourier convolution~\cite{ffc2020}, to enhance performance. However, the implementation of SpikingYOLOX is too cumbersome to be used as an energy-efficient paradigm. Moreover, their effectiveness in specialized, challenging but practical scenarios like aerial imagery~\cite{visdrone,vhr-10} remains underexplored. Driven by this gap, we explore the use of SNNs for aerial object detection, thereby expanding their applicability to this critical domain.

\begin{figure}[!t]
\centering
\includegraphics[width=\columnwidth]{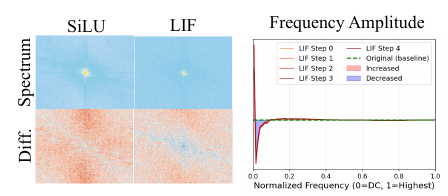} 
\caption[]{The spectrum and frequency distribution of the features in SiLU-based ANN and LIF-based SNN. We also present the difference of the spectrums before and after activation function.}
\label{fig:freq}
\vspace{-0.1in}
\end{figure}

\section{Motivation and Analysis}\label{sec:mot}
\begin{figure*}[!t]
\centering
\includegraphics[width=\textwidth]{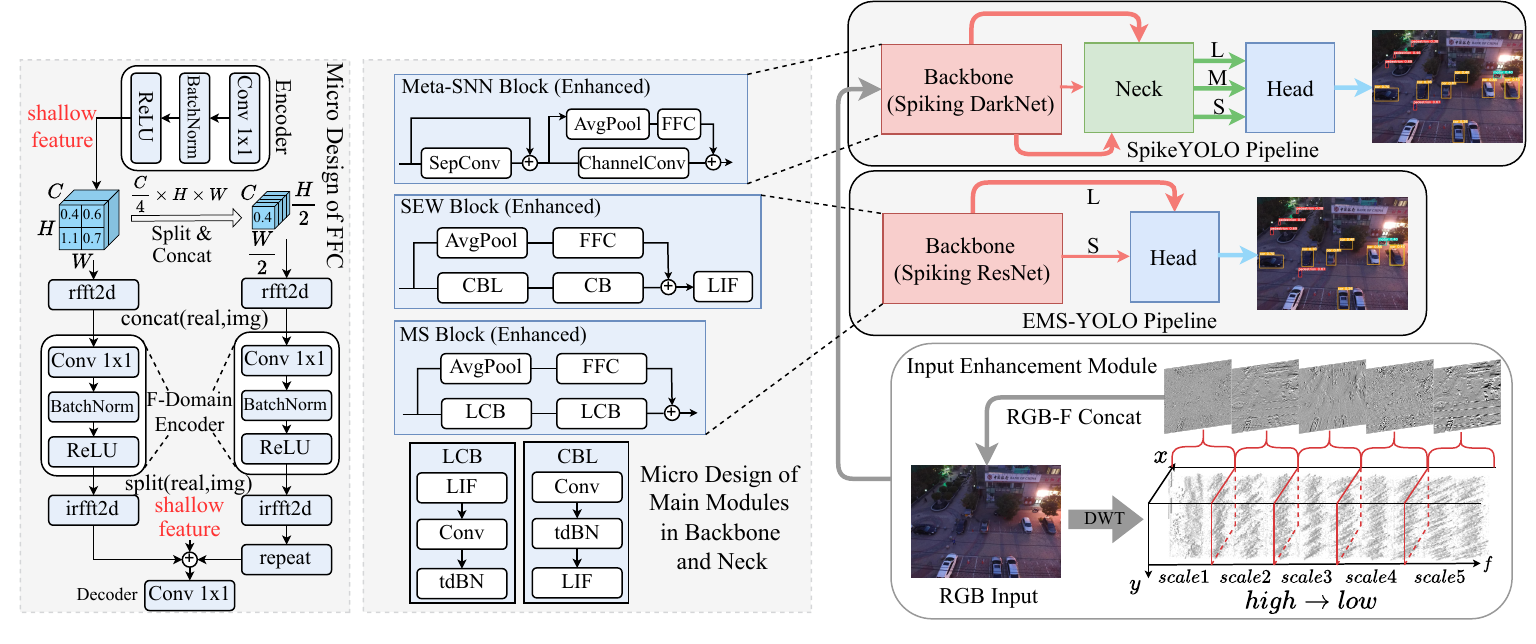} 
\caption[]{Overview of the SNN-based YOLO pipeline and micro design of our proposed modules. The MS block, SEW block, and Meta-SNN block are adopted from MS-ResNet~\cite{ms-resnet}, SEW-ResNet~\cite{sew-resnet}, and SpikeYOLO~\cite{SpikeYOLO}, respectively. Notably, the shortcut paths in the enhanced MS block and SEW block comprise AvgPool and FFC rather than a direct path.}
\label{fig:main}
\end{figure*}
In this section, we conduct detailed theoretical analysis of how spiking neurons influence signal propagation across the network temporally and spatially. All the theoretical analysis of spiking neurons is based on the LIF neuron since it is mostly adopted in SNNs. 
\subsection{Temporal Low-Pass Filter}
The widely used Leaky Integrate-and-Fire (LIF) neuron can be formulated as:
\begin{align}
&V^{-}(t) = (1-\frac{1}{\tau})V(t-1) + \frac{1}{\tau}I(t) \label{eq:lp} \\
&O(t) = \Theta(V^{-}(t)-V_{th}) \label{eq:non-l}\\
&V(t) = V^{-}(t)\cdot(1-O(t)) + V_{reset} \label{eq:reset} \cdot O(t)
\end{align}
where $V^{-}$ and $V$ are the pre- and post-reset potentials, and $\Theta$ is the Heaviside function.
The dynamics of LIF are divided into three parts: linear integration of Eq.(\ref{eq:lp}), non-linear firing of Eq.(\ref{eq:non-l}), and reset of Eq.(\ref{eq:reset}). For the linear integration alone ($V^{-}=V$), applying the Z-transform~\cite{maxformer} gives:
\begin{align}
&V(z) = (1-\frac{1}{\tau}) z^{-1}V(z)+\frac{1}{\tau}I(z) \label{eq:z-trans}\\
&H(z) = \frac{V(z)}{I(z)} = \frac{\frac{1}{\tau}}{1-(1-\frac{1}{\tau}) z^{-1}} \label{eq:h-trans}
\end{align}
Eq.(\ref{eq:h-trans}) is actually the form of a first-order infinite-impulse-response (IIR) low-pass filter~\cite{oppenheim2010} with a single pole $z = 1-\frac{1}{\tau}$. Notably, as $1-\frac{1}{\tau}$, i.e., the pole, approaches 1, LIF neurons exhibit stronger low-frequency selectivity. Accordingly, the decay factor $1-\frac{1}{\tau}$ in Eq.(\ref{eq:lp}) governs the filtering characteristic of the LIF neuron: a larger value makes it more selective for low-frequency components by attenuating high-frequency information more aggressively.

\subsection{Spatial ``Low-Cut'' Filter}
Let us set aside Eq.(\ref{eq:lp}) first as we are trying to analyze effects brought to spatial features of images by LIF. Then, the Heaviside function described by Eq.(\ref{eq:non-l}) is a notable factor:
\begin{align}
    \Theta(x) &= 
    \begin{cases}
        1, & x > 0 \\
        0, & x \le 0 
    \end{cases}
\end{align}
Unlike SiLU or ReLU, such binary activations degrade spatial features in a different way. As mentioned before, we try to find the gap from the perspective of frequency using Fourier Transformation. A standard process of discrete Fourier Transformation (DFT) and the inverse transformation~\cite{oppenheim2010} on each channel $c$ of a raw feature map (or image) $I$ can be formulated as:
\begin{align}
&I^{f}_c(u,v) = \sum_{x=0}^{H-1}\sum_{y=0}^{W-1}I_c(x,y) \cdot e^{-j2\pi(\frac{ux}{H}+\frac{vy}{W})}  \\
&I_c(x,y) = \frac{1}{HW}\sum_{u=0}^{H-1}\sum_{v=0}^{W-1}I_c^f(u,v)\cdot e^{j2\pi(\frac{ux}{H}+\frac{vy}{W})} \label{eq:fft-rev}
\end{align}
where $I^{f}$ is the F-domain signal transformed from $I$. 

Based on Parseval theorem~\cite{oppenheim2010} and Plancherel theorem~\cite{Plancherel}, we can have:
\begin{align}
\sum_{x=0}^{H-1}\sum_{y=0}^{W-1}|I(x,y)|^2 = \frac{1}{HW}\sum_{u=0}^{H-1}\sum_{v=0}^{W-1}|I^f(u,v)|^2 \label{eq:energy-maintain}
\end{align}
The binarized spike activations $g(x,y)=\Theta(I(x,y) - V_{th})$ yield:
\begin{align}
\frac{\sum_{\Omega_{H}}|G(u,v)|^2}{\sum_{\Omega_{L}}|G(u,v)|^2} \propto V_{th} > \frac{\sum_{\Omega_{H}}|I^f(u,v)|^2}{\sum_{\Omega_{L}}|I^f(u,v)|^2} \label{eq:p1}
\end{align}
where $\Omega_H$ and $\Omega_L$ are the empirical regions of high-frequency and low-frequency, respectively. The proofs of Eq.(\ref{eq:energy-maintain}), and Eq.(\ref{eq:p1}) can be found in \textbf{\textit{Appendix~\ref{sec:proof}}}. From Eq.(\ref{eq:p1}), it can be witnessed that the high-frequency energy of $g$ is enlarged, which is mainly attributed to the relative increase of high-frequency components and decrease of low-frequency components according to $V_{th}$. Combined with Eq.(\ref{eq:energy-maintain}), the energy distribution of $g(x,y) = \Theta(I(x,y)-V_{th})$ is altered such that noisy pixel values are amplified, thereby disturbing SNN and degrading performance.

Moreover, to illustrate the correctness of our theoretical analysis, we also present experimental evaluations. Following the most common setting in object detection using SNNs~\cite{ems-yolo} that $V_{th}=0.5$, we present the effect of $\Theta(x)$ caused to the static images in Figure~\ref{fig:freq}. Firstly, both SiLU and Heaviside activation functions enhance the high-frequency components of $I$ since critical features of objects such as edges and contours are highlighted. Meanwhile, $\Theta(x)$ does great harm to the low-frequency components of $I$. Such property is very similar to the high-pass filter that degrades low-frequency components and maintains high-frequency components. However, since the high-frequency is enlarged rather than maintained here, we call this property \textbf{``low-cut''}.  Most intuitively, the loss of spatial low-frequency information uncovers the image noise, which could significantly influence the features of small objects and hinder their recognition. Moreover, features of large objects such as vehicles could be erased, as illustrated in Figure~\ref{fig:quali-freq}. Such results indicate quite different properties of the spiking neuron in the spatial dimension compared to the low-pass property in the temporal dimension. Therefore, solutions targeted at dealing with different properties are essential.

\section{Method}
\subsection{Framework Overview}
Based on the above discussion, we put forward two proposals to compensate information loss brought by the dynamics of LIF in the frequency-domain. As shown in Figure~\ref{fig:main}, we integrate our spatial and temporal modules into two popular SNN-based YOLO pipelines, i.e., EMS-YOLO and SpikeYOLO. Notably, light-weight EMS-YOLO pipeline only uses two scales, i.e., large (L) and small (S), to detect objects, whereas SpikeYOLO uses an additional medium (M) scale. Moreover, EMS-YOLO and SpikeYOLO implement ResNet and DarkNet architectures in spiking version as their backbone, respectively. The input-enhancement module is the first layer of the network, used to enhance inputs. Instead, AvgPool-FFC module is used in the backbone of the pipeline. For SEW Block and MS Block, AvgPool-FFC is inserted into conventional shortcut paths without modules (i.e., direct shortcut paths), while for Meta-SNN block, we only replace the second direct shortcut path with AvgPool-FFC module.

\subsection{Temporal Enhancement with DWT}
We design an input-enhancement module to deal with the temporal low-pass property of LIF neuron. Specifically, this module conducts a frequency-domain transformation on the input static image $I$ to obtain its spectrum first:
\begin{align}
\mathrm{S}(x,y,f) = \mathrm{DWT}(I)
\end{align}
where $\mathrm{DWT}$ denotes the discrete wavelet transformation. Given the nature of $\mathrm{DWT}$, $\mathrm{S}$ can be considered as samplings of a 3-D continuous spectrum, where $\mathrm{S}(x,y,f)$ reflects the magnitude of frequency $f$ at position $(x,y)$ relatively. Given the total time step $T$, we split the interval of $f$ into $T$ scales to obtain $\mathrm{S}(x,y,\mathrm{scale}_i:\mathrm{scale}_{i+1})$, where the frequency from $\mathrm{scale}_{i}$ to $\mathrm{scale_{i+1}}$ decreases. Then, each $\mathrm{S}(x,y,\mathrm{scale}_i:\mathrm{scale}_{i+1})$ is concatenated back to the original RGB image to obtain enhanced RGB-F input. As depicted in Figure~\ref{fig:main}, the visualization of $\mathrm{S}(x,y,\mathrm{scale}_i:\mathrm{scale}_{i+1})$ reflects that the features in the frequency domain contain rich structural details of images, including clear edges of objects and overall structures of the image.

In general, in most SNN-based vision tasks, the input of SNN is created by replicating a static image by $T$ times. In contrast, our temporal enhancement module augments the RGB input with frequency-domain representations, creating variations along the temporal dimension in the F channel. This design offers two key advantages. First, DWT decomposes the image into multiple frequency subbands so that at each spatial location $(x,y)$ along temporal axis, the amount of information to be encoded is significantly reduced. For instance, high-frequency features such as object edges and contours reside only in the high-frequency DWT subband at $t$ and do not appear in low-frequency subbands at other time steps. Consequently, semantically critical features (e.g., edges) only need to be encoded once per spatial location via binary spikes, and these are precisely the features most essential to preserve throughout the network.

Secondly, the input-enhancement module concatenates the frequency subbands $\mathrm{S}$ to the RGB input in order from high to low frequency. This design is driven by the low-pass property of spiking neurons, which inevitably filters out temporal high-frequency information. Crucially, since the neuronal decay factor only takes effect from the second time step onward, the information at the first step is fully preserved. Therefore, placing spatial high-frequency components at early time steps mitigates their degradation. Given the critical role of object edges and contours in detection, the corresponding high-frequency DWT subbands are placed at earlier time steps, while background and low-frequency information are assigned to later ones. We emphasize high-frequency DWT components because they are vital for object detection. This prioritization, however, does not imply that low-frequency information is unimportant; it is a necessary trade-off to mitigate the inevitable loss across all frequency bands. As analyzed in Section~\ref{sec:mot}, the more significant spatial low-frequency loss derives from the Heaviside function. To address that and the issues raised by the tradeoff here, the next section introduces our spatial enhancement module designed to compensate for such loss. 

\subsection{Spatial Compensation with FFC}
As indicated before, the Heaviside activation function tends to enhance high-frequency details, while tending to attenuate important low-frequency information. In the spatial domain, the most intuitive solution is to directly compensate for low-frequency information in the membrane potential of neurons. 

As shown in Figure~\ref{fig:main}, in three kinds of main modules in the model pipeline, we add a new shortcut path to compensate for filtered low-frequency information. Average pooling is first used to narrow the frequency-domain range to low frequencies, preparing low-frequency components of the feature maps. Low-pass filtering also appears before downsampling in anti-aliased CNNs~\cite{antialias2019}. We connect the FFC module behind the average pooling to transform the raw low-frequency features into another space for space-alignment with features in the main path. Moreover, being different from the normal convolution that extracts deep features in the RGB domain, FFC can extract features in the frequency-domain. Finally, the transformed but reserved low-frequency components are added back to the filtered signals. Notably, in SEW-ResNet~\cite{sew-resnet}, it is not valid to directly add floating features from the FFC shortcut to binary spikes. Therefore, we split the CBL block and add the features to the membrane potential before LIF. 

The micro design of FFC is illustrated on the left of Figure~\ref{fig:main}. The first part of FFC is a standard Conv-BN-ReLU block~\cite{batchnorm2015,nair2010}, used as an encoder to obtain the shallow feature map $\mathrm{SF} \in \mathbb{R}^{C \times H \times W} $ of FFC. Then, we choose the first quarter of $\mathrm{SF}$ along the $C$ axis as $\mathrm{SF}[:\frac{C}{4},:,:]$, split it along the $H$ and $W$ axes, and concatenate them to obtain $\mathrm{SF}^{'} \in \mathbb{R}^{C \times \frac{H}{2} \times \frac{W}{2}}$. A real-valued Fast Fourier Transformation (rfft) is applied on $\mathrm{SF}$ and $\mathrm{SF^{'}}$, and the coefficients of real and imaginary parts are concatenated along $C$ to obtain frequency-domain feature maps $\mathrm{FSF}$ and $\mathrm{FSF^{'}}$:
\begin{align}
&\mathrm{CSF}=\mathrm{rfft2d}(\mathrm{SF})\in\mathbb{C}^{C\times H\times K},\\
&\mathrm{FSF}=\mathrm{concat}(\mathrm{CSF.r},\mathrm{CSF.i})\in\mathbb{R}^{2C\times H\times K},\\
&\mathrm{CSF}^{\prime}=\mathrm{rfft2d}(\mathrm{SF}^{\prime})\in\mathbb{C}^{C\times\frac{H}{2}\times K_{\ell}},\\
&\mathrm{FSF}^{\prime}=\mathrm{concat}(\mathrm{CSF}^{\prime}\mathrm{.r},\mathrm{CSF}^{\prime}\mathrm{.i})\in\mathbb{R}^{2C\times\frac{H}{2}\times K_{\ell}}.
\end{align}
Here $K=\lfloor W/2\rfloor+1$ and $K_{\ell}=\lfloor W/4\rfloor+1$.
The utilization of $\mathrm{SF^{'}}$ is to enable multi-scale frequency-domain representation. Rather than downsampling along $H$ and $W$, we obtain a smaller scale $\mathrm{SF^{'}}$ by sampling along $C$, and reshape it to keep consistent with $\mathrm{SF}$ in $C$ dimension. Such a design (i.e., downsampling along $C$) is more suitable for frequency-domain features since rfft is conducted along $H$ and $W$, keeping $H$ and $W$ unchanged causes a smaller loss of frequency information in the original $\mathrm{SF}$.

After the two frequency-domain encoders, the real and imaginary parts of $\mathrm{FSF}$ and $\mathrm{FSF}^{\prime}$ are recombined into complex spectra $\widehat{\mathrm{CSF}}$ and $\widehat{\mathrm{CSF}^{\prime}}$. Inverse rfft with spatial sizes $(H,W)$ and $(H/2,W/2)$ yields:
\begin{align}
&\mathrm{DF}=\mathrm{irfft2d}(\widehat{\mathrm{CSF}})\in\mathbb{R}^{C\times H\times W} \\
&\mathrm{DF}^{\prime}=\mathrm{irfft2d}(\widehat{\mathrm{CSF}^{\prime}})\in\mathbb{R}^{C\times\frac{H}{2}\times\frac{W}{2}}
\end{align}
Finally, $\mathrm{DF}$, repeated $\mathrm{DF^{'}}$, and $\mathrm{SF}$ are added and decoded by a convolution to produce the output of FFC. 

\section{Experiments}
To fully validate the effectiveness of our proposed modules, we conduct experiments on two representative aerial image datasets, i.e., low-altitude VisDrone~\cite{visdrone} and high-altitude VHR-10~\cite{vhr-10}. We report mAP@50(\%) and mAP@50:95(\%) using COCO-style AP metrics~\cite{coco2014}.
\subsection{Experimental Setup}
We use LIF neuron in EMS-YOLO models, where the membrane time constant $\tau$, threshold $V_{th}$ are set to 4 and 0.5, respectively. For SpikeYOLO models, we use I-LIF neuron where the clamping threshold is 4, while time constant $\tau$ follows the same setting as LIF neuron. We mainly train our models on 2 NVIDIA H800 GPUs and adopt the SGD optimizer with learning rate set to $1\times10^{-2}$. All models are trained for 100 epochs on both datasets, and we set batch size of 16 and 12 for VHR-10 and VisDrone, respectively.

\begin{table*}[tbp]
\caption{Results on VisDrone~\protect\cite{visdrone} and VHR-10~\protect\cite{vhr-10}. ``+Ours'' uses the AvgPool-FFC module and four-channel RGB-F input. The time step $T$ for Backbone-Head and Backbone-Neck-Head architecture is set as 5 and 1, respectively. The theoretical energy consumption (EC) is obtained using the theory proposed in~\protect\cite{mac-ac,horowitz2014}. FPS denotes frames per second. For each baseline/+Ours pair, higher mAP values are shown in bold. Moreover, the energy consumption comparison with ANNs can be found in \textbf{\textit{Appendices~\ref{sec:energy} and~\ref{sec:ann}}}.}
\centering
\resizebox{0.9\textwidth}{!}{
\begin{tabular}{cccccrrcrr}
\hline
\multicolumn{1}{c}{\multirow{2}{*}{Architecture}} & \multicolumn{1}{c}{\multirow{2}{*}{Model}} & \multirow{2}{*}{Params} & \multicolumn{1}{c}{\multirow{2}{*}{FPS}} & \multicolumn{3}{c}{VisDrone} & \multicolumn{3}{c}{VHR-10} \\ \cline{5-10} 
\multicolumn{1}{c}{} & \multicolumn{1}{c}{} &  &  & \begin{tabular}[c]{@{}c@{}}EC \\ (mJ)\end{tabular} & \begin{tabular}[c]{@{}c@{}}mAP@\\ 50(\%)\end{tabular} & \begin{tabular}[c]{@{}c@{}}mAP@\\ 50:95(\%)\end{tabular} & \multicolumn{1}{c}{\begin{tabular}[c]{@{}c@{}}EC\\  (mJ)\end{tabular}} & \begin{tabular}[c]{@{}c@{}}mAP@\\ 50(\%)\end{tabular} & \begin{tabular}[c]{@{}c@{}}mAP@\\ 50:95(\%)\end{tabular} \\ \hline
\multirow{10}{*}{EMS-YOLO} & SEW-ResNet18 & 9.58M & 4.15 & 11.6 & 12.2 & 5.2 & 11.2 & 60.6 & 29.7 \\
 & +Ours & 11.8M & 3.80 & 16.1 & \textbf{13.1} & \textbf{5.4} & 15.7 & \textbf{60.8} & \textbf{30.0} \\ \cline{2-10} 
 & MS-ResNet18 & 9.65M & 4.38 & 12.3 & 13.7 & 5.7 & 12.4 & 59.3 & 26.8 \\
 & +Ours & 10.03M & 3.28 & 18.1 & \textbf{14.8} & \textbf{6.2} & 18.2 & \textbf{59.8} & \textbf{28.4} \\ \cline{2-10} 
 & EMS-ResNet18 & 9.36M & 3.93 & 11.8 & 13.9 & 5.8 & 11.6 & 61.3 & 26.9 \\
 & +Ours & 9.91M & 3.55 & 12.5 & \textbf{15.1} & \textbf{6.2} & 12.0 & \textbf{63.3} & \textbf{29.8} \\ \cline{2-10} 
 & MS-ResNet34 & 15.1M & 2.87 & 16.2 & 16.6 & 7.3 & 15.9 & 63.5 & 32.0 \\
 & +Ours & 15.8M & 2.15 & 18.4 & \textbf{17.5} & \textbf{7.8} & 17.9 & \textbf{66.2} & \textbf{32.6} \\ \cline{2-10} 
 & EMS-ResNet34 & 14.4M & 3.37 & 14.6 & 15.9 & 7.0 & 13.2 & 63.6 & 30.4 \\
 & +Ours & 14.9M & 2.75 & 15.8 & \textbf{17.4} & \textbf{7.7} & 15.0 & \textbf{65.9} & \textbf{30.8} \\ \hline
\multirow{6}{*}{SpikeYOLO} & SpikeYOLOv8l & 65.4M & 3.00 & 88.6 & 37.4 & 21.9 & 87.8 & 55.3 & 31.0 \\
 & +Ours & 74.1M & 2.73 & 90.2 & \textbf{38.7} & \textbf{22.7} & 89.6 & \textbf{59.6} & \textbf{35.1} \\ \cline{2-10} 
 & SpikeYOLOv8m & 45.7M & 3.57 & 66.2 & 35.7 & 20.3 & 65.9 & 58.3 & 34.7 \\
 & +Ours & 51.7M & 3.56 & 69.5 & \textbf{36.9} & \textbf{20.7} & 69.2 & \textbf{60.4} & \textbf{35.1} \\ \cline{2-10} 
 & SpikeYOLOv8s & 23.1M & 6.20 & 35.3 & 34.1 & 19.8 & 33.2 & 59.4 & 34.2 \\
 & +Ours & 25.9M & 6.12 & 37.6 & \textbf{35.3} & \textbf{20.4} & 36.4 & \textbf{61.3} & \textbf{35.7} \\ \hline
\end{tabular}
}
\label{tab:main}
\end{table*}

\subsection{Overall Performance}
\textbf{Quantitative Evaluation.} The main quantitative results are presented in Table~\ref{tab:main}. Overall, our modules increase the parameters of the baselines by fewer than 10\% on average and effectively improve the accuracy of all baselines. On the VisDrone dataset, our enhanced EMS-YOLO baselines achieve an average improvement of 1.1~pp in mAP50. For larger backbone-neck-head models, the modules deliver a further gain, averaging 1.2~pp. The improvement is more pronounced on the VHR-10 dataset, where enhanced SpikeYOLO models attain average gains of 2.8~pp in mAP50 and 2.0~pp in mAP50:95, respectively. Intriguingly, on the VisDrone dataset, the enhanced SpikeYOLOv8s model, with only 25.9M parameters, outperforms the 45.7M-parameter 
\begin{figure}[!t]
\centering
\includegraphics[width=\columnwidth]{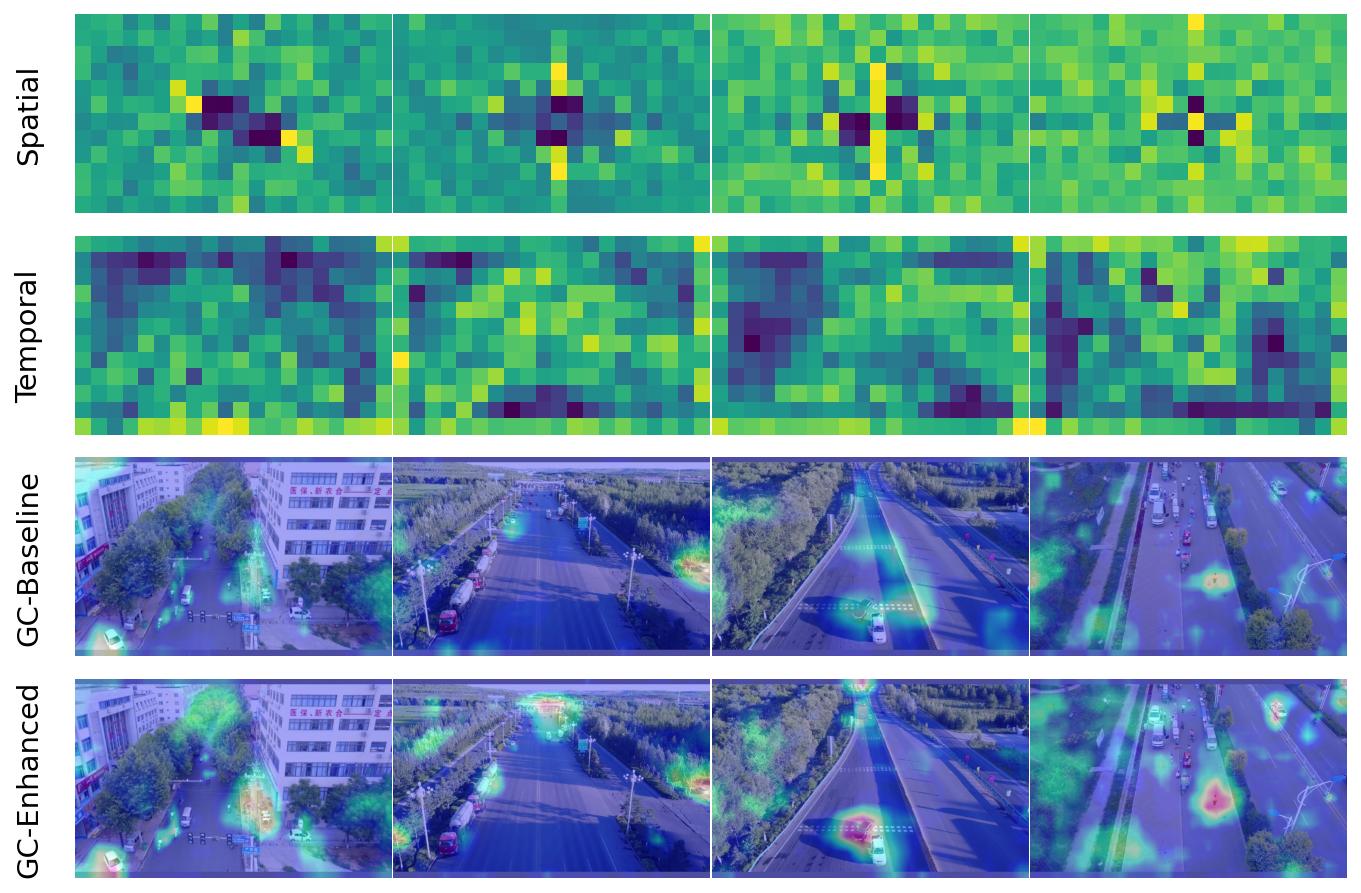} 
\caption[]{Spectrum and GradCAM~\cite{gradcam} visualizations of the backbone features. We present differences (baseline - enhanced) of spectrum to intuitively present the gap.
}
\label{fig:quali-freq}
\vspace{-0.1in}
\end{figure}
SpikeYOLOv8m in terms of mAP50:95 and achieves competitive mAP50 performance. Moreover, on VHR-10, the enhanced SpikeYOLOv8s achieves the highest mAP50 and mAP50:95 among the SpikeYOLO models in Table~\ref{tab:main}. With 25.9M parameters, it outperforms the larger enhanced v8m and v8l models, indicating that increasing model size alone does not necessarily improve accuracy on this dataset. In contrast, our input-enhancement module extracts frequency information from static images and injects it into the temporal input sequence. This design effectively exploits the temporal dynamics unique to SNNs. Spatially, by aligning with the spatial characteristics of SNNs, our FFC module compensates for the attenuated low-frequency components. It achieves comparable or even superior results with only a minimal increase in parameters. Moreover, averaged over the 16 model--dataset pairs in Table~\ref{tab:main}, theoretical energy consumption increases by 16.3\%, preserving the inherent efficiency of SNNs. Importantly, latency tests on the widely-used NVIDIA Jetson AGX platform for UAVs show minimal additional delay for SpikeYOLO pipeline, confirming practical viability for edge deployment. 

\begin{figure*}[!t]
\centering
\includegraphics[width=\textwidth]{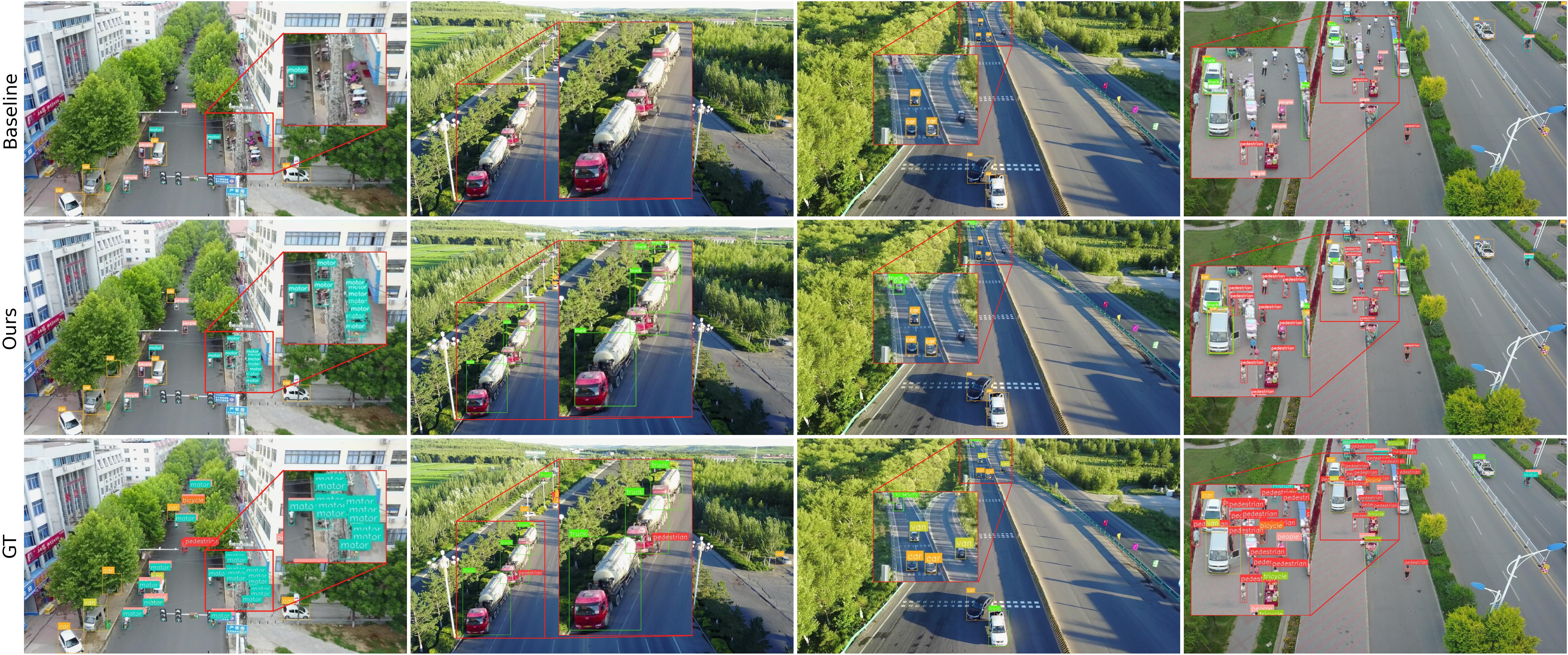} 
\caption[]{Object detection results on the VisDrone Dataset. We present the results of EMS-YOLO in our main paper. The baseline we use is the EMS-YOLO model using EMS-ResNet34 as backbone. The enhanced baseline presents better ability to recognize small objects. For instance, in the third column, our modules enable the baseline to detect two tiny trucks (green label) at the edge of the image. Due to paper limitation, more results of different baselines and datasets are presented in supplemental materials.
}
\label{fig:quali-visdrone}
\end{figure*}

\noindent\textbf{Qualitative Evaluation.}
We show object detection results on the test set, illustrated in Figure~\ref{fig:quali-visdrone}. It can be witnessed that our method significantly improves the baseline's ability to locate and recognize small objects. Notably, the baseline fails to detect large-area trucks (second column, Figure~\ref{fig:quali-visdrone}). This is unexpected, as large objects are typically easier to detect, and thus demands explanation. Following the analysis in our methodology, we present frequency-domain and GradCAM~\cite{gradcam} visualizations to explore the reasons. As depicted in Figure~\ref{fig:quali-freq}, in the spatial domain, the baseline lacks low-frequency components relative to the enhanced version, whereas in the temporal domain, it is deficient in high-frequency components. A critical loss of spatial low-frequency information in the second column explains the detection failure. Objects like the large white trucks are dominated by low-frequency signals; their loss therefore prevents detection, as indicated before.
For the third and fourth samples in Figure~\ref{fig:quali-freq}, despite significantly amplifying high-frequency energy, the baseline fails to detect small objects. This implies a tendency in SNNs to misclassify fine-grained object features as noise, resulting in missed detections.

The temporal spectrums show a reverse pattern compared to the spatial spectrum. While the baseline is adversely affected by the temporal low-pass property of neurons, losing important high-frequency encodings, our input-enhancement module that is positioned at the first layer mitigates this issue and maintains such information through the network, as evidenced in deep layers like EMS-ResNet34’s last layer. Finally, GradCAM visualizations confirm that the enhanced model focuses more sharply on target objects, whereas the baseline shows weak or no attention, as most information has been filtered out both temporally and spatially.

\subsection{Ablation Study}
To further explore the nature of SNNs in the frequency domain, we conduct an ablation study to evaluate more detailed impacts brought by our proposed modules.
\begingroup
\setlength{\intextsep}{0pt}
\setlength{\abovecaptionskip}{0pt}
\begin{figure}[H]
\centering
\includegraphics[width=\columnwidth]{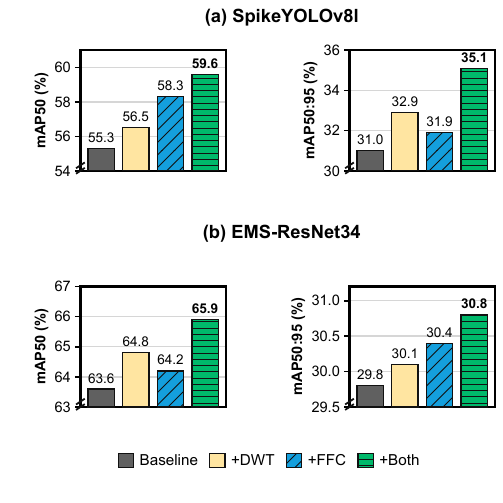}
\caption{Ablation study on VHR-10 with SpikeYOLOv8l and EMS-ResNet34. ``+DWT'' uses the input-enhancement module, ``+FFC'' uses the spatial-enhancement module, and ``+Both'' uses both modules. Bars show absolute mAP50 and mAP50:95 values, with the highest value in each group shown in bold.}
\label{fig:abl-vhr}
\end{figure}
\endgroup
\begin{figure}[!t]
\centering
\includegraphics[width=0.8\columnwidth]{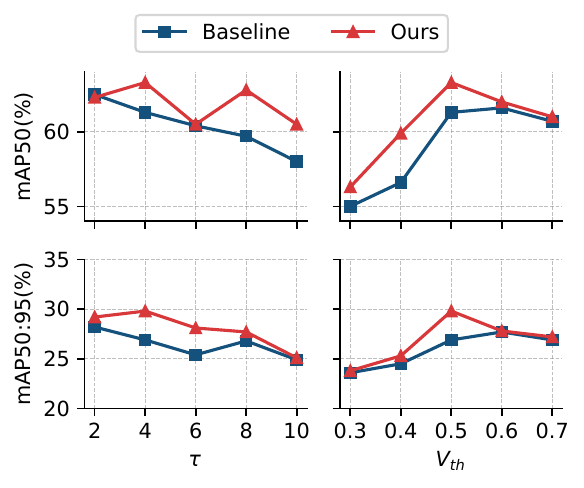}
\vspace{-0.2in}
\caption[]{Effect of time constant $\tau$ and threshold $V_{th}$ on the EMS-ResNet18 baseline and the enhanced model on VHR-10.}
\label{fig:hyper}
\vspace{-0.1in}
\end{figure}
As shown in Figure~\ref{fig:abl-vhr}, both of our proposed modules enhance the baseline performance on VHR-10. Used individually, the DWT and FFC modules improve both mAP50 and mAP50:95 for SpikeYOLOv8l and EMS-ResNet34. When combined, they yield significantly greater gains, surpassing the sum of their individual contributions. Such results illustrate that in SNNs, spatial low-frequency information and temporal high-frequency information are complementary and can influence each other. The combination of DWT and FFC fully exploits the spatial-temporal nature of SNNs, thus obtaining a synergistic performance gain.

Moreover, as discussed in Section~\ref{sec:mot}, the time constant $\tau$ and the threshold $V_{th}$ are accordingly related to temporal and spatial properties of SNNs. Therefore, we perform a sensitivity analysis to evaluate the robustness of our modules against extreme scenarios. As presented in Figure~\ref{fig:hyper}, as $\tau$ increases and $1-\frac{1}{\tau}$ approaches 1 so that low-frequency selectivity of spiking neurons grows, the enhanced model achieves significantly higher mAP@50 and mAP@50:95 compared to the original baseline. This demonstrates our method's robustness in extreme temporal low-pass scenarios. For the results of $V_{th}$, we find that as $V_{th}$ increases, the gap between the baseline and the enhanced version is becoming minimal. Such results illustrate that fewer spatial low-frequency components degrade when the threshold is larger. Moreover, the performance of both the baseline and our enhanced model improves with an increasing threshold. This observation is seemingly counterintuitive, as a lower threshold typically allows more spikes to be emitted, which one might expect to facilitate more complete information encoding. Intriguingly, this phenomenon is explained by our theoretical analysis and corroborated by experimental results, validating our proposals.

\section{Conclusion}
In this paper, we established a frequency-domain analysis of spiking neurons, identifying their temporal low-pass filtering and spatial ``low-cut'' filtering as key factors limiting SNN performance in aerial object detection tasks. To address this, we propose SNN-STEM with spatial and temporal enhancement modules. Experimental results on aerial image datasets demonstrate that these modules effectively improve the detection accuracy of SNN-based frameworks while preserving their efficiency, providing a novel frequency-aware perspective for enhancing SNN-based vision systems, particularly for challenging object detection tasks.

\section*{LLM Usage Statement}
We used ChatGPT (OpenAI) for language polishing. It was not involved in the research conception or design and had no substantive impact on the core scientific contributions. The authors reviewed all edits and take full responsibility for the manuscript.

{
\small
\bibliographystyle{ieeenat_fullname}
\bibliography{ijcai26}
}
\clearpage
\appendix
\setcounter{figure}{0}
\setcounter{table}{0}
\setcounter{equation}{0}
\renewcommand{\thefigure}{S\arabic{figure}}
\renewcommand{\thetable}{S\arabic{table}}
\renewcommand{\theequation}{S\arabic{equation}}
\renewcommand{\theHfigure}{supp.\arabic{figure}}
\renewcommand{\theHtable}{supp.\arabic{table}}
\renewcommand{\theHequation}{supp.\arabic{equation}}
\twocolumn[{\centering\Large\bfseries Supplementary Material\par\vspace{1em}}]
\section{Theoretical Energy Consumptions}
\label{sec:energy}
\subsection{SNN energy (spike-driven AC-based)}
For an SNN, spiking synaptic layers predominantly perform spike-triggered accumulation (AC) operations. Let $T_{\mathrm{SNN}}$ denote the number of model simulation time steps used to process each input image. We use $T_{\mathrm{SNN}}=T$, with $T$ reported for each SNN model in Table~\ref{tab:ann}. Let $\gamma_\ell(\tau)\in[0,1]$ denote the mean presynaptic spike rate that feeds the synapses of layer $\ell$, measured as the dataset average of binary spikes per neuron per timestep. Concretely, if layer $\ell$ is driven by spikes from layer $\ell-1$, we set
\begin{equation}
\gamma_\ell(\tau):=r_{\mathrm{spk}}^{(\ell-1)}(\tau),
\end{equation}
where $r_{\mathrm{spk}}^{(\cdot)}$ is defined as the spatial firing rate. Following the approximation in prior work, the number of synaptic operations (SOPs) at layer $\ell$ is estimated by
\begin{equation}
\mathrm{SOPs}_\ell=T_{\mathrm{SNN}}\cdot\gamma_\ell(\tau)\cdot\mathrm{FLOPs}_\ell,
\end{equation}
where $\mathrm{FLOPs}_\ell$ here denotes the MAC count of the corresponding dense layer (i.e., the cost if computed densely as in an ANN). Then the theoretical compute energy of a spiking synaptic layer is
\begin{equation}
E_\ell^{\mathrm{SNN}}=E_{\mathrm{AC}}\cdot\mathrm{SOPs}_\ell.
\end{equation}

\subsection{First-layer convention and total \#OP}
In many algorithm papers, all SNN layers are treated as AC-only. In our setting, the input is real-valued; therefore the first layer typically involves dense MACs. Moreover, since the FFC module we used mainly involves MACs, we compute $E^{\mathrm{SNN}}$ as follows:
\begin{equation}
E^{\mathrm{SNN}}=\sum_{\ell\in\mathcal{S}}E_{\mathrm{AC}}\cdot\mathrm{SOPs}_\ell
+\sum_{\ell\in\mathrm{FFC},s_1}E_{\mathrm{MAC}}\cdot\mathrm{FLOPs}_l
\end{equation}
where $L$ is the number of layers and $\mathcal{S}$ is the set of general spiking layers. Following~\cite{mac-ac}, we use $E_{\mathrm{MAC}}=4.6\,\mathrm{pJ}$ and $E_{\mathrm{AC}}=0.9\,\mathrm{pJ}$ to estimate computation energy for 32-bit floating-point operations in 45\,nm technology.

\section{Proofs of ``Low-Cut'' of Spiking Neurons}
\label{sec:proof}
Given raw features $I(x,y)$, their counterpart $I^f(u,v)$ after Discrete Fourier Transformation (DFT) and the reverse transformation are formulated as:
\begin{equation}
I^f(u,v)=\sum_{x=0}^{H-1}\sum_{y=0}^{W-1}I(x,y)\cdot e^{-j2\pi(\frac{ux}{H}+\frac{vy}{W})}
\end{equation}
\begin{equation}
I_c(x,y)=\frac{1}{HW}\sum_{u=0}^{H-1}\sum_{v=0}^{W-1}I_c^f(u,v)\cdot e^{j2\pi(\frac{ux}{H}+\frac{vy}{W})}
\end{equation}
Then, the image-domain energy of $I(x,y)$ equals:
\begin{align}
\sum_{x,y}|I(x,y)|^2&=\sum_{x,y}I(x,y)\cdot I^*(x,y)\\
&=\sum_{x,y}\left[\frac{1}{HW}\sum_{u,v}I^f(u,v)e^{j2\pi(\frac{ux}{H}+\frac{vy}{W})}\right]\\
&\quad\cdot\left[\frac{1}{HW}\sum_{u',v'}(I^f)^*(u',v')e^{-j2\pi(\frac{u'x}{H}+\frac{v'y}{W})}\right]\\
&=\frac{1}{(HW)^2}\sum_{u,v}\sum_{u',v'}I^f(u,v)(I^f)^*(u',v')\\
&\quad\cdot\sum_{x,y}e^{j2\pi(\frac{(u-u')x}{H}+\frac{(v-v')y}{W})}
\end{align}
where $I^*(x,y)$ and $(I^f)^*$ are the complex conjugate functions of $I$ and $I^f$, respectively.

\begin{lemma}
For $k\in\mathbb{Z}$ and positive integer $H$, the discrete exponential functions exhibit:
\begin{equation}
\sum_{x=0}^{H-1}e^{j2\pi\frac{kx}{H}}=
\begin{cases}H,&k=mH,\ m\in\mathbb{Z}\\0,&k\ne mH\end{cases}
\end{equation}
\end{lemma}
The derivation of Lemma~1 is as follows: if $k=mH$, then, based on Euler's formula, $e^{j2\pi mx}=1$, thus:
\begin{equation}
\sum_{x=0}^{H-1}e^{j2\pi\frac{kx}{H}}=H
\end{equation}
Otherwise, if $k\ne mH$, denote $e^{j2\pi\frac{k}{H}}$ by $r$. Then, we have:
\begin{align}
\sum_{x=0}^{H-1}e^{j2\pi\frac{kx}{H}}&=\sum_{x=0}^{H-1}r^x\\
&=\frac{1-r^H}{1-r}=\frac{1-e^{j2\pi k}}{1-e^{j2\pi k/H}}
\end{align}
Recall that $e^{j2\pi k}=1$. Thus, we obtain the sum in Lemma~1 as 0. Based on Lemma~1, we can get:
\begin{align}
&\sum_{x,y}e^{j2\pi(\frac{(u-u')x}{H}+\frac{(v-v')y}{W})}\\
&=\left(\sum_{x=0}^{H-1}e^{j2\pi\frac{\Delta u x}{H}}\right)\left(\sum_{y=0}^{W-1}e^{j2\pi\frac{\Delta v y}{W}}\right)\\
&=\begin{cases}HW,&u=u',\ v=v'\\0,&\text{otherwise}\end{cases}\\
&=H\cdot W\cdot\delta_{u,u'}\cdot\delta_{v,v'}
\end{align}
where $\Delta u=u-u'$ and $\Delta v=v-v'$, and $\delta$ is the Kronecker delta function. Finally, we can derive:
\begin{align}
&\sum_{x=0}^{H-1}\sum_{y=0}^{W-1}|I(x,y)|^2\\
&=\frac{1}{(HW)^2}\sum_{u=0}^{H-1}\sum_{v=0}^{W-1}I^f(u,v)\\
&\quad\cdot(I^f)^*(u,v)\cdot(H\cdot W)\\
&=\frac{1}{HW}\sum_{u=0}^{H-1}\sum_{v=0}^{W-1}|I^f(u,v)|^2
\label{eq:parseval}
\end{align}

\begin{figure}[t]
\centering
\includegraphics[width=0.92\linewidth]{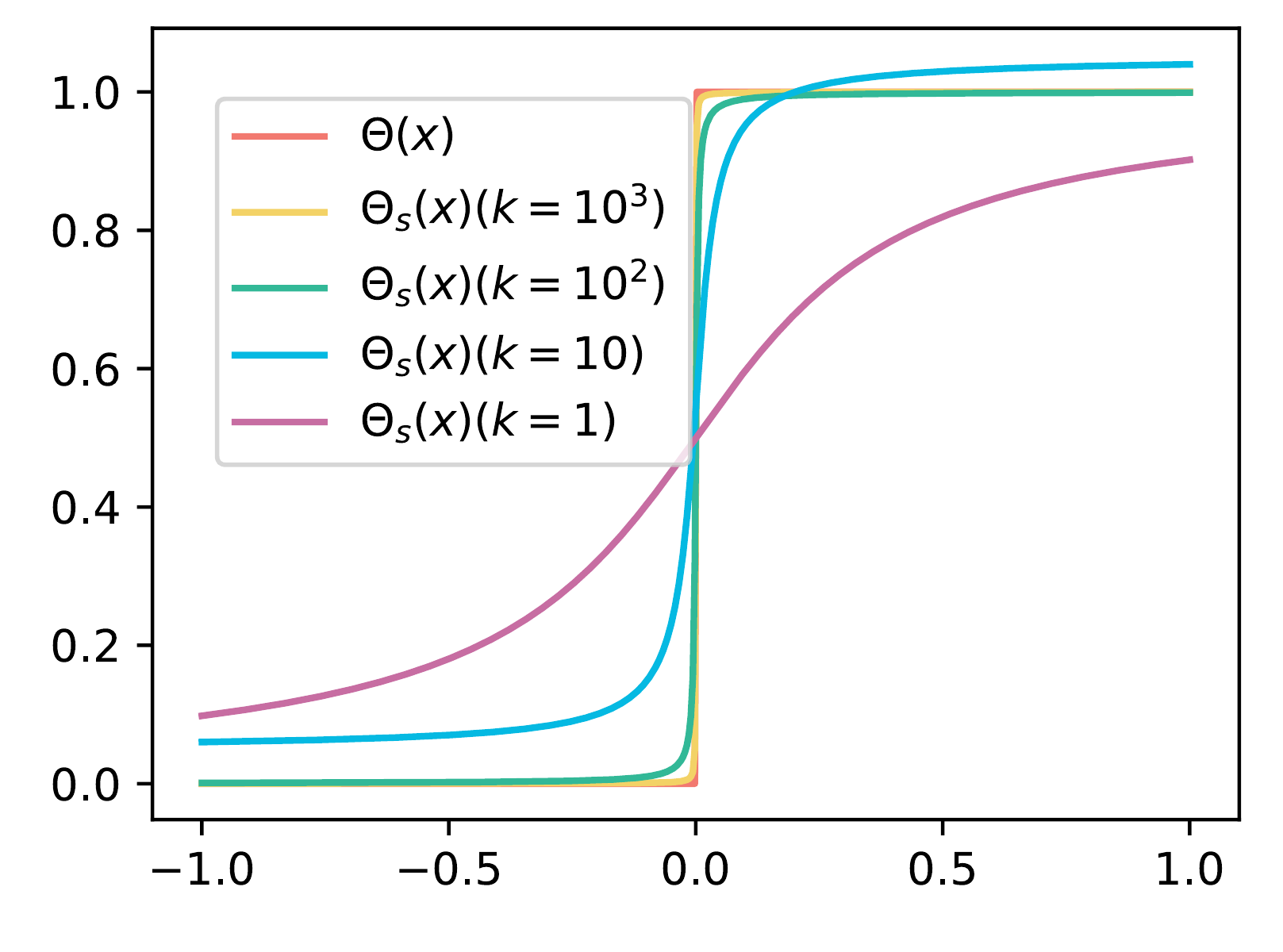}
\caption{Function Curves of $\frac{1}{\pi}\arctan(k\pi x)+0.5$ and $\Theta$.}
\label{fig:surrogate}
\end{figure}

Next, we start to analyze the effects of the Heaviside function $\Theta$ on spatial information. Notably, since $\Theta$ is non-linear and non-differentiable, it is extremely hard to directly analyze its frequency-domain effect on $I(x,y)$. Therefore, we use surrogate function $\Theta_s=\frac{1}{\pi}\arctan(k\pi x)+0.5$, whose effect on the $I(x,y)$ is very close to the effect of $\Theta$ on the $I(x,y)$, as presented in Figure~\ref{fig:surrogate}, to simplify our derivation.

\begin{table*}[t]
\centering
\caption{Overall evaluation on VHR-10 dataset~\cite{vhr-10}. We conduct thorough evaluations on both accuracy and energy consumption (EC). Here $T$ denotes the number of model simulation time steps per input image, and a dash indicates that this quantity is not applicable to ANN models.}
\label{tab:ann}
\small
\setlength{\tabcolsep}{7pt}
\begin{tabular}{lrrrrr}
\toprule
Methods & Params (M) & $T$ & EC (mJ) & mAP50 (\%) & mAP50:95 (\%)\\
\midrule
Faster-RCNN &43.2&--&390.72&67.3&35.2\\
YOLOv3 &103.7&--&1428.8&58.2&33.0\\
YOLOv5m &25.1&--&295.05&60.3&34.3\\
YOLOv8m &25.9&--&279.40&63.2&35.5\\
YOLOv11m &20.1&--&301.56&62.7&36.7\\
\midrule
EMS-ResNet18+ours &9.91&5&12.0&63.3&29.8\\
MS-ResNet18+ours &10.03&5&18.2&59.8&28.4\\
SEW-ResNet18+ours &11.8&5&15.7&60.8&30.0\\
EMS-ResNet34+ours &14.9&5&15.0&65.9&30.8\\
MS-ResNet34+ours &15.8&5&17.9&66.2&32.6\\
SpikeYOLOv8m+ours &51.7&1&69.2&60.4&35.1\\
\bottomrule
\end{tabular}
\end{table*}

For simplicity, we assume that $V_{\mathrm{th}}$ has been subtracted from $I(x,y)$. Let $\mu=\langle I\rangle$ denote the mean of $I(x,y)$, and let $g(x,y)=\Theta_s(I(x,y))$. For $k>0$, $\Theta_s$ is smooth on the real line. Taylor's theorem gives
\begin{align}
g(x,y)&=\Theta_s(\mu)+\Theta_s'(\mu)(I-\mu)\\
&\quad+\frac{1}{2}\Theta_s''(\mu)(I-\mu)^2\notag\\
&\quad+\frac{1}{6}\Theta_s'''(\mu)(I-\mu)^3+R_4(x,y),
\end{align}
where $R_4(x,y)=\Theta_s^{(4)}(\xi_{x,y})(I(x,y)-\mu)^4/4!$ for some $\xi_{x,y}$ between $\mu$ and $I(x,y)$.
We denote $\tilde I=I-\mu$, let $N=HW$, and write $F=\mathcal{F}\{\tilde I\}=\tilde I^f$. Taking the unnormalized forward DFT gives
\begin{align}
G(u,v)&=NC_0\delta_{u,0}\delta_{v,0}+C_1F(u,v)\\
&\quad+\frac{C_2}{N}(F\circledast F)(u,v)\notag\\
&\quad+\frac{C_3}{N^2}(F\circledast F\circledast F)(u,v)\notag\\
&\quad+\widehat{R_4}(u,v).
\end{align}
Here $C_0=\Theta_s(\mu)$, $C_1=\Theta_s'(\mu)$, $C_2=\frac{1}{2}\Theta_s''(\mu)$, $C_3=\frac{1}{6}\Theta_s'''(\mu)$, and $\widehat{R_4}=\mathcal{F}\{R_4\}$. The operator $\circledast$ denotes two-dimensional circular convolution:
\begin{align}
(A\circledast B)(u,v)&=\sum_{a=0}^{H-1}\sum_{b=0}^{W-1}A(a,b)\\
&\quad\cdot B((u-a)\bmod H,(v-b)\bmod W).
\end{align}
Generally, it is not that possible that $\mu=0$ so that the original frequency spectrum $\tilde I^f$ will be scaled by $C_1=\Theta_s'(\mu)=\frac{k}{1+(k\pi\mu)^2}<1$. Therefore, we can find that all frequency components in the original spectrum are degraded. Let $J(x,y)$ denote the features before threshold shifting, so that $I(x,y)=J(x,y)-V_{\mathrm{th}}$ and $\mu=\langle J\rangle-V_{\mathrm{th}}$. For fixed $J$, the threshold therefore modulates the linear spectral gain as $C_1(V_{\mathrm{th}})=k/[1+(k\pi(\langle J\rangle-V_{\mathrm{th}}))^2]$.

For the remaining non-linear terms of $\tilde I^f$ such as $\mathcal{F}\{\tilde I^2\}$, since $\mathcal{F}\{\tilde I^2\}=N^{-1}(F\circledast F)$, they create new frequency components. If $\tilde I^f$ has components with frequency $(u_1,v_1)$ and $(u_2,v_2)$, the new frequencies created in $\mathcal{F}\{\tilde I^n\}$ $(n\ge2)$ include:
\begin{align}
&(0,0),\\
&(2u_1,2v_1),\\
&(2u_2,2v_2),\\
&(u_1+u_2,v_1+v_2),\\
&(u_1-u_2,v_1-v_2)
\end{align}
Most of those new components have higher frequencies than the original components. Therefore, although there will be new low-frequency components compensating for the loss in the original spectrum, more unexpected higher-frequency information can be added, relatively degrading low-frequency information of the output $g(x,y)$. Moreover, combined with Eq.~\eqref{eq:parseval}, the entire energy of raw spatial image $I(x,y)$ is significantly degraded. However, high-frequency noise can become relatively more prominent, thus disturbing and degrading SNNs' performance.

\section{Comparison with ANNs}
\label{sec:ann}
We conduct an accuracy and energy consumption comparison between SNNs enhanced with our modules and some widely-used industrial ANNs (e.g., YOLOv3~\cite{yolov3_2018}, YOLOv5, YOLOv8, YOLOv11, and Faster-RCNN~\cite{fasterrcnn2015}). We believe these models are representative since they are widely used on edge devices such as UAVs for object detection. All SNN models are trained on the VHR-10 dataset~\cite{vhr-10} for 100 epochs with a batch size of 16 using one NVIDIA R6000 GPU. ANN models are trained for 250 epochs using the toolkit provided by the Ultralytics\footnote{\url{https://github.com/ultralytics/ultralytics}} project. As shown in Table~\ref{tab:ann}, SNNs achieve significant energy efficiency while maintaining comparable detection accuracy. For instance, the enhanced EMS-ResNet18 with only 9.91M parameters outperforms YOLOv11m and YOLOv8m in terms of mAP50 and achieves comparable accuracy in terms of mAP50:95, while enabling $25.1\times$ and $23.3\times$ energy efficiency, respectively. We believe such efficiency can provide insights about how to enhance the endurance of UAVs in future industrial applications. More qualitative results can be found below.

\onecolumn
\section{Additional Qualitative Results}
\label{sec:qualitative}
\begin{figure}[!ht]
\centering
\includegraphics[width=\linewidth]{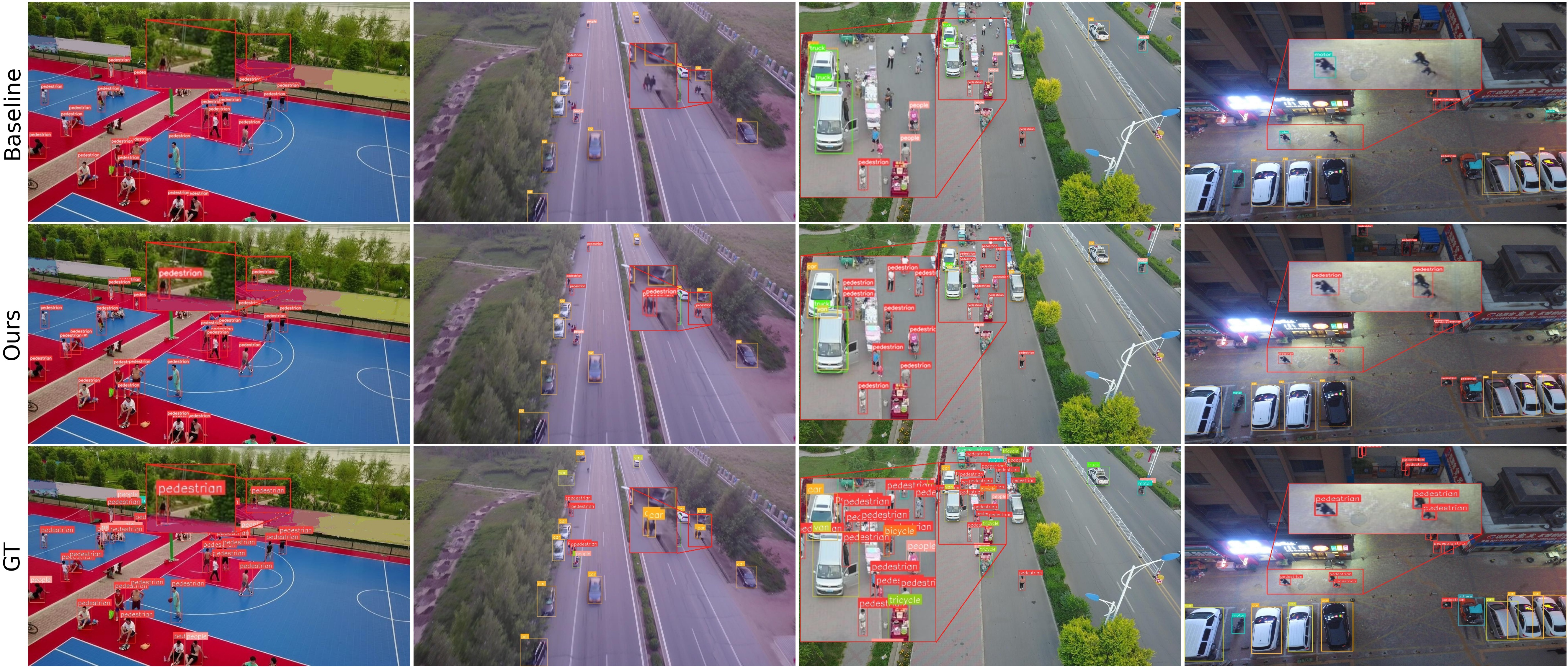}
\caption{Object detection results of MS-ResNet34 and the enhanced model on the VisDrone Dataset. Notably, for the second column, two pedestrians are mislabeled as cars (dataset error), while our method still helps the baseline to correctly detect them.}
\end{figure}
\begin{figure}[!ht]
\centering
\includegraphics[width=\linewidth]{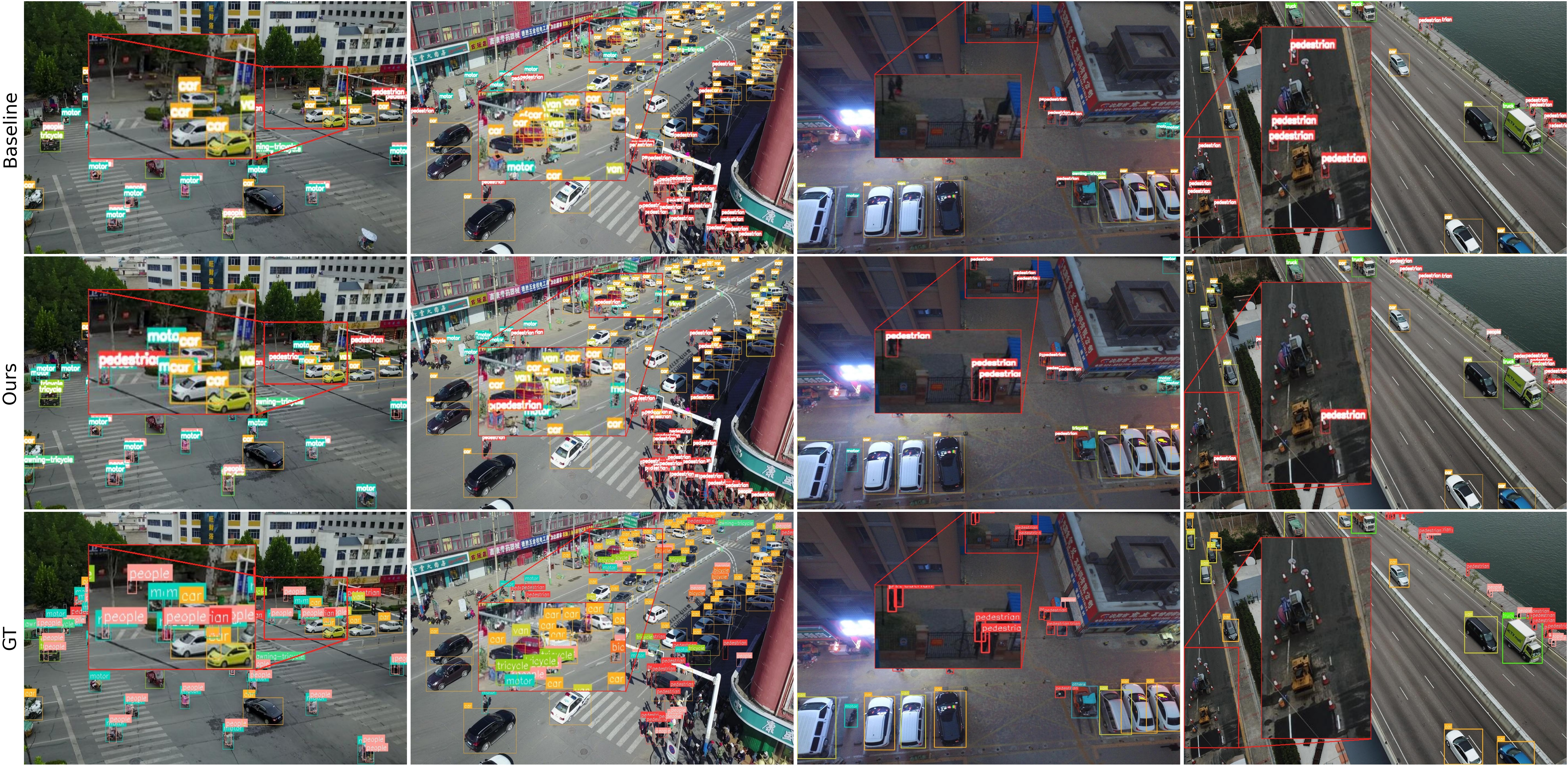}
\caption{Object detection results of SpikeYOLOv8l and the enhanced version on the VisDrone Dataset.}
\end{figure}
\clearpage
\begin{figure}[!ht]
\centering
\includegraphics[width=\linewidth]{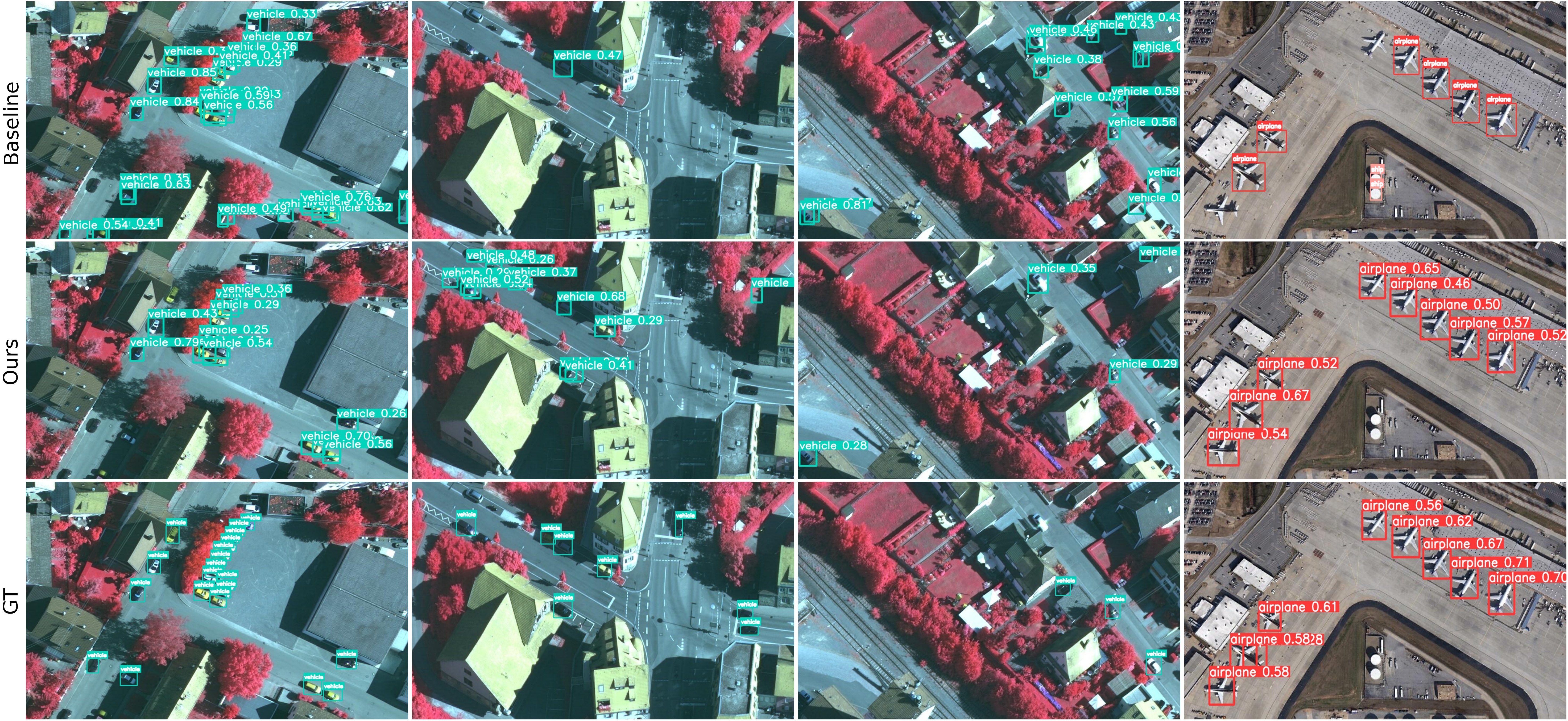}
\caption{Object detection results of EMS-ResNet18 and the enhanced version on the VHR-10 Dataset.}
\end{figure}
\begin{figure}[!ht]
\centering
\includegraphics[width=\linewidth]{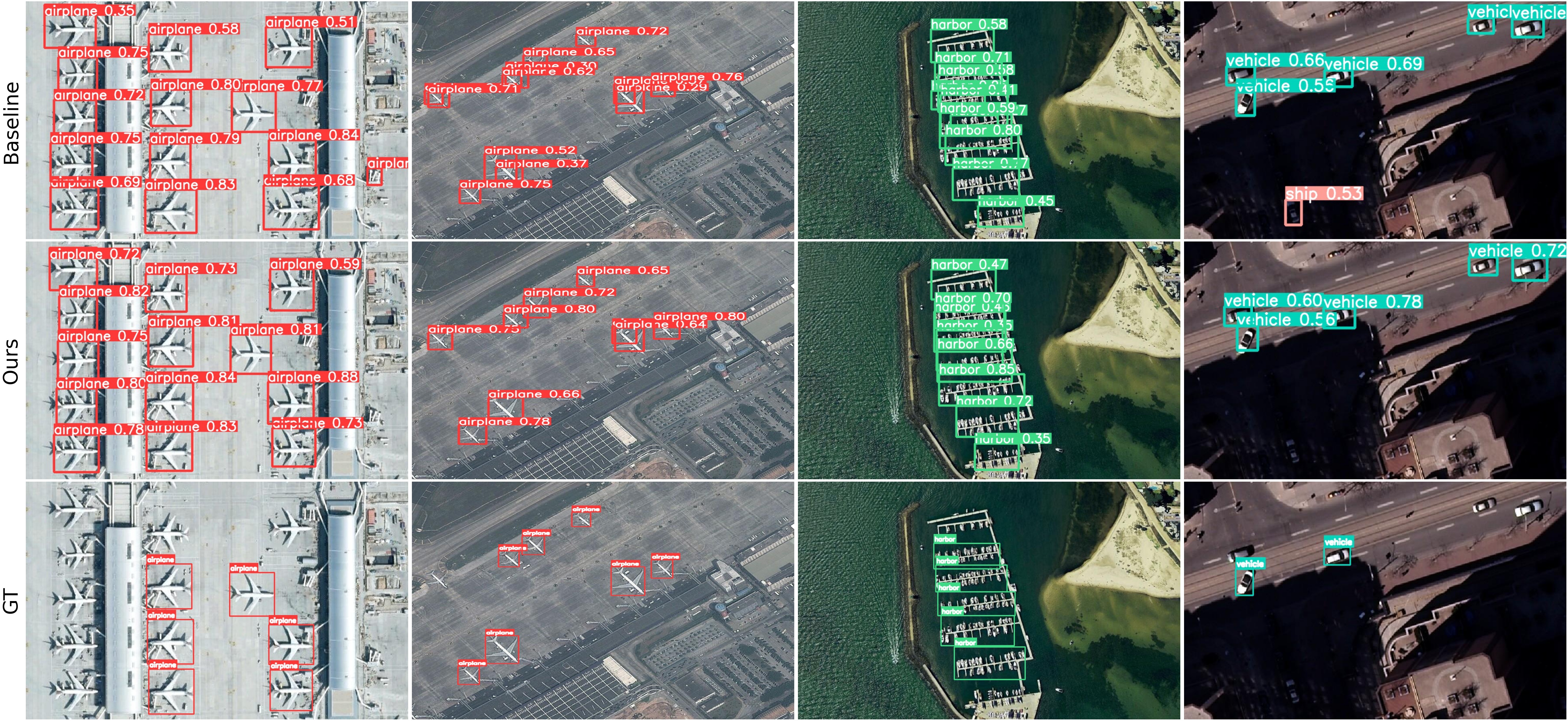}
\caption{Object detection results of SEW-ResNet18 and the enhanced version on the VHR-10 Dataset.}
\end{figure}
\clearpage
\begin{figure}[!ht]
\centering
\includegraphics[width=\linewidth]{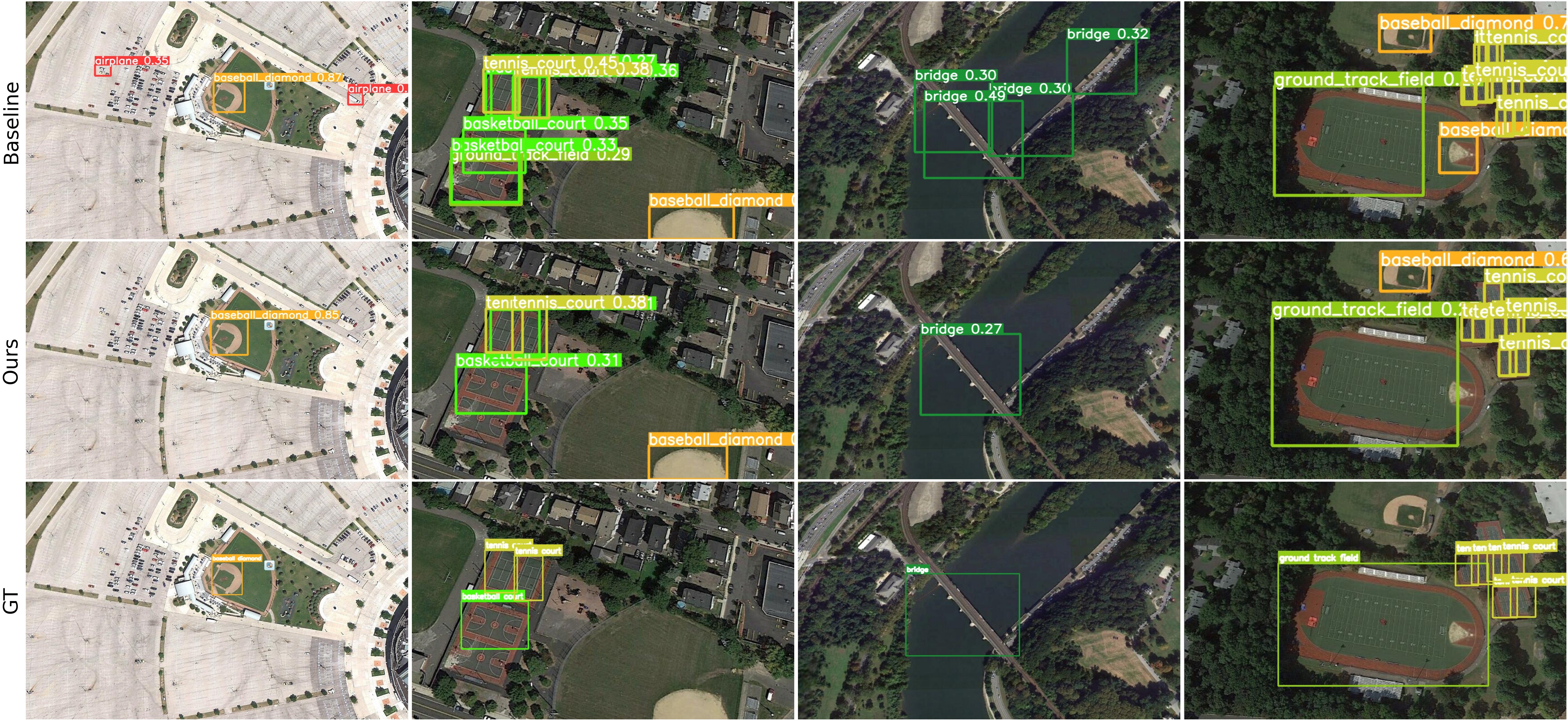}
\caption{Object detection results of MS-ResNet18 and the enhanced version on the VHR-10 Dataset.}
\end{figure}
\begin{figure}[!ht]
\centering
\includegraphics[width=\linewidth]{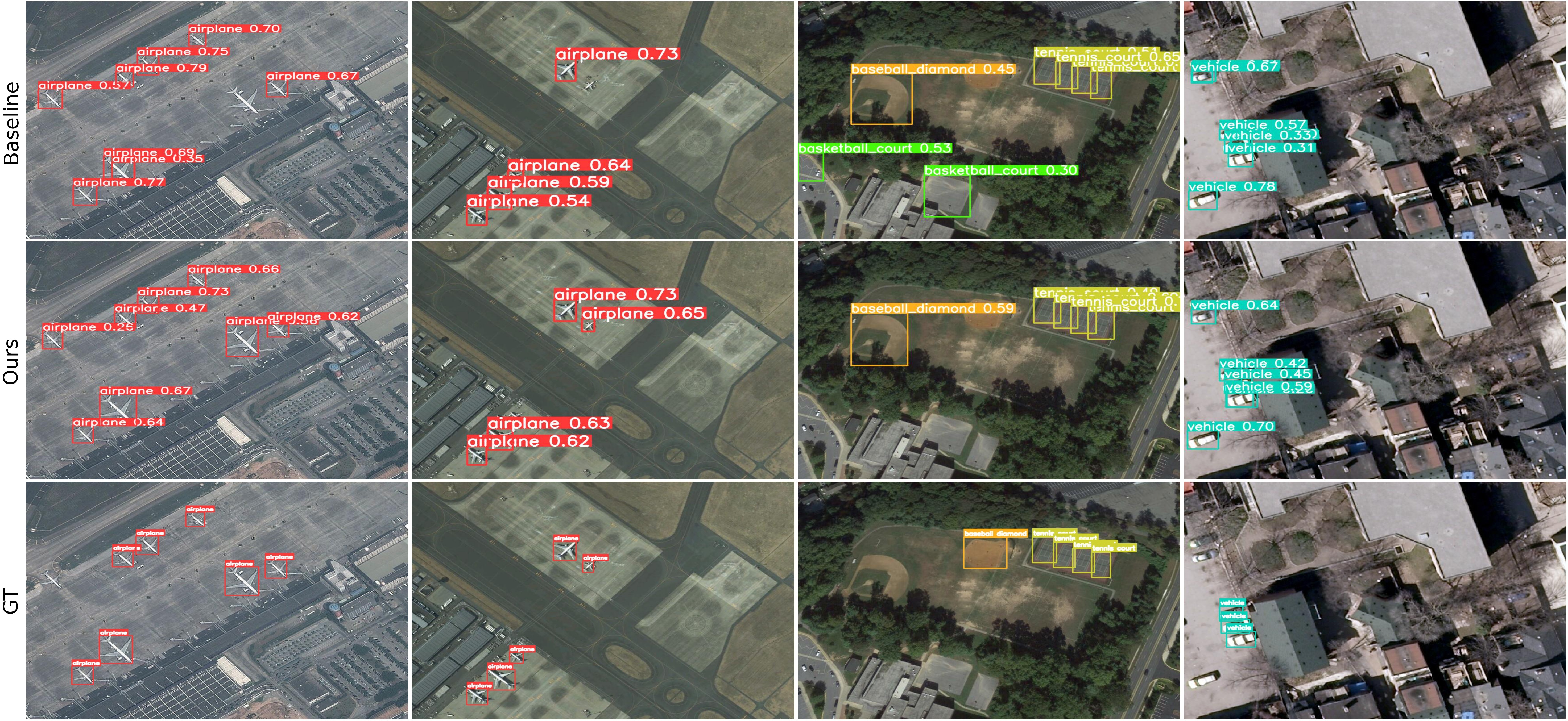}
\caption{Object detection results of SpikeYOLOv8m and the enhanced version on the VHR-10 Dataset.}
\end{figure}
\clearpage
\begin{figure}[!ht]
\centering
\includegraphics[width=\linewidth]{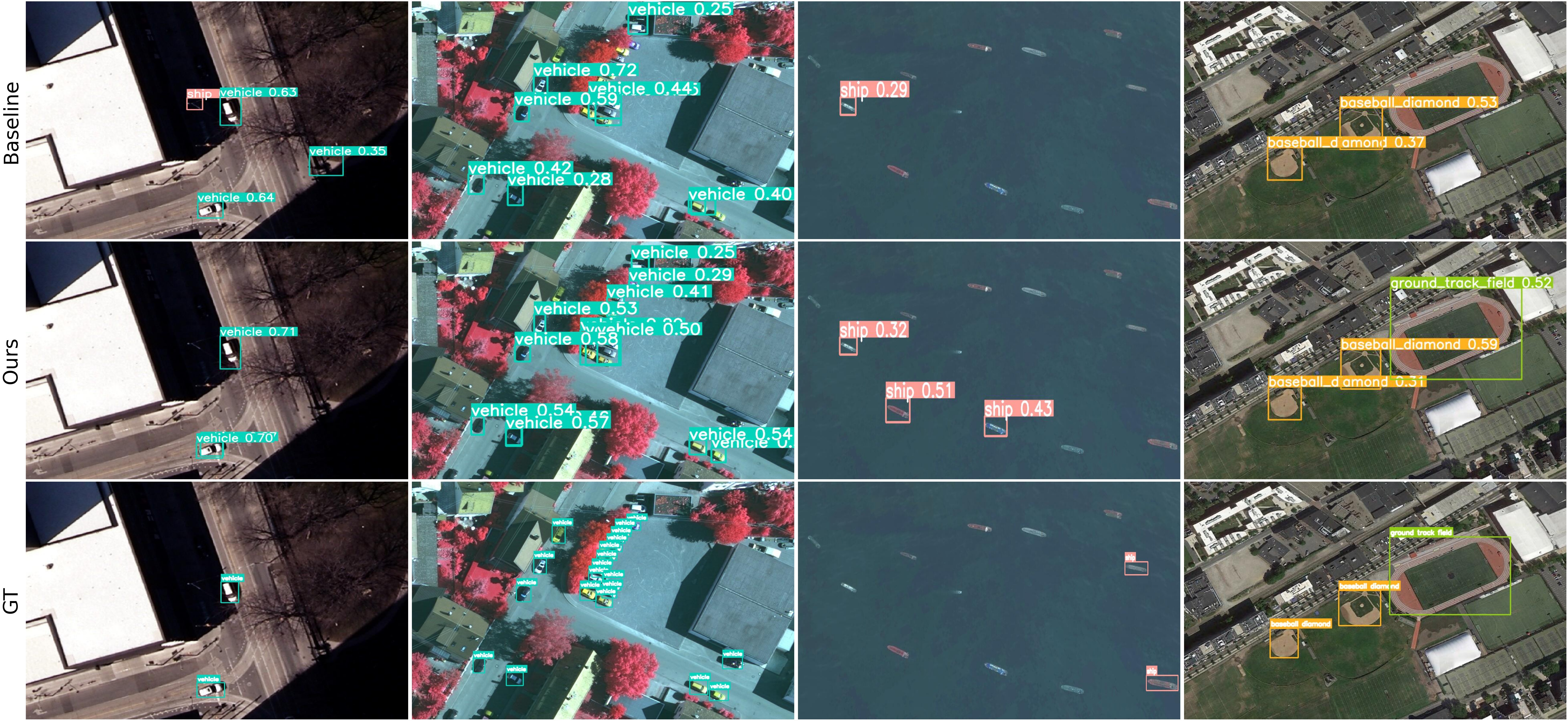}
\caption{Object detection results of SpikeYOLOv8s and the enhanced version on the VHR-10 Dataset.}
\end{figure}

\end{document}